# A New Oscillating-Error Technique for Classifiers

Kieran Greer, Distributed Computing Systems, Belfast, UK.
kgreer@distributedcomputingsystems.co.uk.
Version 1.5c

***Abstract –*** This paper describes a new method for reducing the error in a classifier. It uses an error correction update that includes the very simple rule of either adding or subtracting the error adjustment, based on whether the variable value is currently larger or smaller than the desired value. While a traditional neuron would sum the inputs together and then apply a function to the total, this new method can change the function decision for each input value. This gives added flexibility to the convergence procedure, where through a series of transpositions, variables that are far away can continue towards the desired value, whereas variables that are originally much closer can oscillate from one side to the other. Tests show that the method can successfully classify some benchmark datasets. It can also work in a batch mode, with reduced training times and can be used as part of a neural network architecture. Some comparisons with an earlier wave shape paper are also made.

## 1  Introduction

Neural networks and classifiers in general are statistical processors. They all work by trying to reduce the error in the system through an error correction method that includes transposition through a function. Neural networks in particular, are based loosely on the human brain, with a distributed architecture of relatively simple processing units. Each neural unit solves a small part of the problem, where collectively, they are able to solve the whole problem. Being statistical classifiers, they try to converge to some solution without any level of intelligence outside of the pre-defined function. This works very well for a statistical system, but the simulation of a brain-like neuron could include a little bit more. It does get involved in different kinds of biochemical reaction [4][29] and may even have a type of memory [26]. For this paper, the neuron is able to react to its input and apply a very simple rule of either adding or subtracting the error adjustment, based on whether the variable value is currently larger or smaller than the desired value, and on a variable by variable basis. The decision is based on the most basic of reactions and so it could be part of an automatic theory. It is also well known that resonance is a feature of real brain operations and other simulation models [3][14]. The idea of resonance would be to use the data shape to determine what values go together, where earlier research [13] and this paper suggest that the data shape can be represented by a single averaged value. The procedure is shown to work surprisingly well and be very flexible and so it should be taken seriously as a general mechanism.

The rest of this paper is organised as follows: section 2 briefly outlines the reasons for the new method. Section 3 introduces some related work and section 4 describes the theory behind the new classifier. Section 5 runs through a very simple test example, while section 6 gives the result of some tests on real datasets. Finally, section 7 gives some conclusions to the work.

## 2   Reasons for the New Method

The proposed method would give the component slightly more flexibility, or if arguing for a neural component, then a small amount of intelligence, but still keep it at a most basic and automatic level. Each variable can reduce its error in a way that best suits it, with a dampening effect that is independent of the other variables. Basically, if the data point (variable value) is less than the desired value, the weight adjustment is added to it and if it is larger than the desired value, the weight adjustment is subtracted from it. This means that variables of the same input set to the neuron could be treated differently when the neuron applies the function, which gives added flexibility to the convergence procedure. Through a series of transpositions or levels in the classifier, a variable that is far from the correct value can be adjusted by the full amount in the same direction each time. A variable that is at the correct value can oscillate around it and therefore some of the adjustment size can even be removed. The method is implemented here in matrix form, but as it uses a neuron-like architecture, it can be compared more closely with neural networks, or simply as a general update mechanism. The weight correction can also be added or subtracted and not multiplied, where the data works best with some form of normalisation, but considering a binary-style of reduction, it does not take many steps for the error to reduce. The error correction is also calculated by using the input and desired output values only and not any intermediary error value sets. Although, this maybe considers the whole matrix to be a single hidden unit. One other advantage of the method is the fact that it is not necessary to fine-tune the classifier, with appropriate random weight sets, for example. The weight correction procedure will always be the same and only a stopping criterion is required, along with the dataset pre-processing.

## 3   Related Work

Related work would therefore include neural networks [27][31] and the resonance type in particular [3][14]. The Adaptive Resonance Theory is an example of trying to use resonance, created by a matching agreement, as part of a neural network model. It is also categorical in nature, but can learn category patterns and includes a long-term memory component that is a matrix of weight updates. The primary intuition behind the ART model is that object identification and recognition generally occur as a result of the interaction of 'top-down' observer expectations with 'bottom-up' sensory information and the idea of resonance is the agreement between these two processes. Resonance suggests a repeating value or state, which then suggests an averaged value, which is why it may be possible to represent a wave shape that way. The Fuzzy-ART system uses what is called a one-shot learning process, where each input item can be categorised after just one presentation. Cellular automata possibly have some relation as well [32][5], because the new neural component is at a similar level of complexity. It is not usual for a neural component to make a decision, but the decision is so simple that it might be compared to a reaction. The paper [15] is also interesting in this respect, with their Gauss-Newton gradient descent Marquardt algorithm. It uses batch processing to compute the average sum of squares over the dataset error, and can add or subtract a value from the step value, which is also a feature of the related Marquardt-Levenberg algorithm. So in fact, these algorithms do make a similar decision, although it applies to the weight rather than the value itself. The rule that the new neuron uses can probably make the best fit result non-linear, even if it is linear with respect to time.

Attempts to optimise the learning process have been made since the early days of neural networks. Kolmgorov's theorem [2][22] is often used to support the idea that a neural network can successfully define any arbitrary function using just one hidden layer [17]. While Deep Learning has improved on this, it would be an idea of the model of this paper. The theorem states that each multivariate continuous real-valued function can be represented as a superposition and composition of continuous functions of only one variable. The paper [10] gives a summary of some early attempts, including batch processing and even the inclusion of rules, but as part of different types of learning frameworks. It is interesting that rules and discrete categories or activations, are all quite old ideas. More recently, the deep learning neural network models [18] adopt a policy of many more levels than the earlier backpropagation ones. These new networks include a feedback from one level to previous ones, as well as continuously refining the function, to learn mid-level structures or features. Some Convolutional Neural Networks can also be trained in a one-shot mode. The paper [19], for example, can train the network using only one labelled example per category, as part of a data reduction or transformation process. One-shot learning therefore appears to be the term that was originally used. The paper [12] also uses batch processing or averaging of the input dataset, and uses the term single-pass to mean a similar thing.

Resonance is mentioned because an earlier neural network paper [13] tried to encapsulate the dataset shape into a single averaged value and these papers [3][12] that are interested in resonance also try to condense the input data rows into vectors of single averaged values. In that case, a relative size of a scalar becomes important, but discriminating comparisons must still be made. To help with this, the dataset is separated for each output category, so that the averaged value applies to one category only. The justification is that each neuron always has to accommodate all of the data that passes through it and so it has to produce an average evaluation for that. Thus, averaging the input data could become a very cheap way of describing the data shape. While the closest classifier might be a neural network, this new model uses a matrix-like structure that contains a number of transitions from one layer to the next. These are however relatively simple transformations of adding or subtracting a value and are really just steps in the same error reduction procedure.

## 4   Background Theory and Method Description

The theory of the new mechanism started with looking at the wave shape paper [13], which is described first with some new details. After that, the new oscillating error mechanism is described.

### 4.1   Wave Shape Algorithm

This was proposed in [13] as an alternative way of looking at the relative input and output value sets. The idea was that the value differences would describe a type of wave shape and similar shapes could be combined in the synapses, as they would produce the same type of resonance. That design also uses average values, where both the input and the output can be summed and averaged over each column (all data rows), to represent each variable field with the average value. Tests do in fact show a substantial reduction in the error of the average input to the average output using this method and even on established datasets, such as the Wine dataset [7][28]. The problem was that while the error could be reduced, it was reduced to an average output value that is not very accurate for each specific instance. For example, if the output values are 1, 2 and 3, then the input dataset could be averaged to produce a value close to 2, but this is not very helpful when trying to achieve an exact output value of 1 or 3. That procedure, based strongly on shape, could be more useful for modelling the synapses, whereas the neuron needs to compare with the desired result. Therefore, using actual values instead of differences is probably more appropriate. For example, if the input dataset is 2, 8, 4, 5, 10; then you can measure the average of these values, or the average of their differences: 6, -4, 1, 5. As part of a theory, the synapses could consider shape more than an actual value, as they try to sync with each other, while the neuron compares with the actual result. So possibly, modelling the network can consider that neurons and synapses are measuring a different type of quantity over the same value set and for a different purpose – one to reinforce a type of signal (synapse) and one to produce a more exact output (neuron). As stated however, averaging over the whole dataset makes the network too general and so possibly the ideas of the next section can be tried.

### 4.2   Oscillating-Error Method

This is the new algorithm of the paper and resulted from trying to make the input to output mapping of the last section more accurate. The new neuron can take an input from each variable or column and adjust it by either adding or subtracting the weight update, on a variable by variable basis. As the error oscillates from one side to the other, a bit of it gets removed, as the current difference and so it will necessarily reduce in size. The new neuron is therefore the same as a traditional one, except for the inclusion of the rule as part of the calculation and separate weight sets for each category, during training. The new mechanism has been tried using batch values, as for section 4.1, but the learning procedure is different to the earlier models mentioned in section 3. It has been implemented in a matrix form of levels that pass each input to the next level and is not as a flexible neural network, but the units that are used would be suitable for neural networks in general. The calculations are really only the ones described later and the equations suggest that time would be linear with increasing dataset size or number of levels. The tested datasets required only a second or less to be classified, where additional time to create the initial category groupings might be the only consideration. The pre-processing however creates the batch rows, only 1 for each category and so much fewer row numbers are subsequently used for training.

This paper only considers categorical data, where each input row belongs to a single category. If represented by a single output neuron however, this can still produce a range of output values, but they represent a discrete set instead of a continuous one. In the case of the Wine dataset [7], the 3 output categories can be represented by the values 0, 0.5 and 1.0, for example. As described in section 4.1, the current wave shape method is not accurate enough, as it averages over all categories. The new method therefore sums and averages over each category group separately. In effect, it divides the dataset into batches, representing the rows in each category and produces an averaged data row for each category group. For the Wine dataset, there are therefore three sets of input data, one for each category, represented by 3 averaged data rows. These then update the classifier separately, which stores different sets of weight or error correction values for each category group. The weight value sets can then be combined into a single weight value set after they are learned, to be used over any new input. For the Wine dataset, during training for example, the structure would store 3 sets of 13 weight or error correction values, relating to the 3 output categories and the 13 input variables. After the error corrections have been determined, the 3 values for each variable are summed and averaged to produce the value to be used by the classifier on any classification task. This also becomes the starting set of weight update values for the next network layer. The method also vertically adjusts the error, instead of using a multiplication factor.

### 4.3   Training Algorithm

The following algorithm helps to describe the process:

1. Group all data rows for each output category. Each group is then processed separately during training.
    a. For each category group, sum and average all input points for each variable (or data column) to produce an averaged data row for that category.
2. To train the classifier:
    a. Pass each data row of group values through the layers and update for the new layer.
        i. For the input layer, present each averaged data row to the classifier.
        ii. For other layers, present the last set of weight adjusted inputs.
    b. For the current layer, create the new weight correction set as follows:
        i. If the value is smaller than the desired output value, then add the previous layer's averaged weight correction value to it.
        ii. If the value is larger than the desired output value then subtract the previous layer's averaged weight correction value from it.
        iii. Measure the difference between the new weight-corrected value and the desired category output. Take the absolute value of that as the weight error correction value for the data point in the category group.
        iv. The error value can also be summed and compared with earlier layers, to evaluate the stopping criterion.
    c. The weight update method is essentially a single event that sets the value for the category group in the layer.
    d. After evaluating the weight sets for each category group separately, average over them and store the averaged list as a new transposition layer in the matrix.

3. The transposed values can also be stored as each new layer is added, to make the next learning phase quicker. It can continue from the last layer, instead of running the values through the whole matrix again.
4. Go to step 2 to create the next matrix layer in the structure, and repeat the process until a stopping criterion is met.
5. A stopping criterion can be number of iterations, or if the total error does not reduce by a substantial amount anymore.

During training, each layer creates a set of error correction weights for each of the output categories. After training, these weight sets are then summed and averaged to produce a final set for that layer. At the end of the process, there is then a matrix-like structure of layers, each with a single set of error correction values, one for each input variable. Any new input data row can be passed through each layer and the related correction value added or subtracted from it using the simple rule. This produces an output value for each variable (column) in the data row. The final layer is a single neuron that represents the discrete output categories. All of the input values can be summed and averaged to produce an exact output value. If a margin of error is allowed, then the closest category group can be selected.

The strength of the process lies in the fact that input values that are very far from the desired one can continue to move towards it, while ones that are closer can start to oscillate around it and do not need to be moved away by the same error correction[1]. This gives added flexibility to the learning process and makes the variables a bit independent of each other. This is therefore a very simple idea, with a minimum of disturbance to the mechanical and automatic nature of the traditional neuron. The following equation *Equ. 1* can be used to determine the variable value at a level in the classifier. This is used by the classifier after it has learned the transposition layers' weights and therefore only needs to adjust the input values using these weights. Equation *Equ. 2* describes the error correction rule and fits into *Equ. 1* as the $X_{ij}$ or the network value for variable *j* at level *i*.

---

[1] For example, in a standard neural network: if point 1 has an error of 10 and point 2 has an error of 0, then if you subtract 10 from both to correct point 1, the point 2 error actually increases to 10.

$$X = (\sum_{i=1}^{m} \sum_{j=1}^{n}(Xij)) / n \qquad \text{Equ. 1.}$$

Where:

$X_{ij} = X_{i-1j} + EC_{ij}$ if $X_j <= O$ and  Equ. 2.

$X_{ij} = X_{i-1j} - EC_{ij}$ if $X_j > O$.

Where:

O = desired output value.
X = final output value.
$X_{ij}$ = input value for variable (column) j after transposition in matrix layer i.
$EC_{ij}$ = error correction for variable j in layer i.
n = total number of variables.
m = total number of matrix layers.

## 5   Example Trace of a Scenario

The following scenario traces through the process for a dataset with 5 variables. The example assumes that they have already been grouped for the output category and is intended to demonstrate the error correction procedure only. The desired output category value is '4'. The following steps show how the variables can converge to that value at each iterative step[2].

*Averaged Input row values to layer 1:   3, 8, 5, 10, 2*

*Output category value:   4*

*Input-Output Differences =   Abs(4 – 3), Abs(4 – 8) , Abs(4 – 5) , Abs(4 – 10) , Abs(4 – 2)*

*Absolute error =   1, 4, 1, 6, 2*

- Next iteration: take the input values and adjust, by adding or subtracting the error correction.

---

[2] If there is more than one output category value, then the weight values for each group can conflict and the error might not automatically reduce to 0, as is this example. That is also why the categories are grouped separately for training.

- For variable 1, for example: 3 is less than 4, so add 1 to it. For variable 2: 8 is larger than 4, so subtract 4 from it, and so on.
- Determine the new difference from the desired output to get the new weight set.

*Input plus/minus error correction to layer 2: 4, 4, 4, 4, 4*
*Input-Output Differences =  Abs(4 – 4), Abs(4 – 4) , Abs(4 – 4) , Abs(4 – 4) , Abs(4 – 4)*
*Absolute error =   0, 0, 0, 0, 0*

Continue until the stopping criterion is met. In this case, the error is now 0. It is interesting that with a single output category, this method reduces the error to 0 in 1 step. If there are several output categories and their weights sets are averaged, then the weight update will not necessarily reduce the error to 0. Also, if there was another layer, then it would adjust input values that are '0, 0, 0, 0, 0' and not the original input value set.

## 6   Test Results

A test program has been written in the C# .Net language. It can read in a data file, normalise it, generate the classifier from it and measure how many categories it subsequently evaluates correctly. The classifier was designed with only one output node, as described in section 4.2. The input values were also normalised. Therefore, 3 categories would produce desired output values of 0, 0.5 and 1. The conversion from a category to a real number is not implicit in the data and so it is possible to use a value range to represent each category, just as easily as a single value. It might be interesting however for numerical data, if specific output values can be learned accurately. The error margin that is discussed as part of the result does not relate to distributions, but relates to the smallest margin around the output value representing the category that will give the best percentage of correct classifications. The representative value is still what the classifier tries to learn, but then a value range round that can only reduce the number of errors. For example, consider 3 categories again. These are represented by the output values 0 (category 1), 0.5 (category 2) and 1.0 (category 3), which gives a gap of '0.5' between each value. It would therefore be possible to measure up to 49% of that gap, either side of a category value and still be 100% reliable with respect to the category classification. A 20% error margin, for example, would be calculated as 0.5 * 20 / 100 = 0.1. This would mean that a range of 0.4 – 0.6 would be classified as the category 2 and anything outside of this range could be classified as incorrect. A 15% margin of error would mean that the range would have to be 0.425 – 5.75, and so on. So a smaller error margin would simply indicate that the classifier could be more accurate to an exact real value and there is no ambiguity over the results presented in this paper. Binary data could also be handled equally easily.

The process is completely deterministic. There are no random variables and so a dataset with the same parameter set will always produce the same result. Two types of result were measured. The first was an average error for each row in the dataset, after the classifier was trained, calculated as the average difference between actual output and the desired output value. The second measurement was how many categories were correctly classified, but also with a consideration of the value range (error margin) just discussed. If increasing the margin around a category value did not substantially increase the number of correct classifications, then maybe it would not be worthwhile.

### 6.1   Benchmark Datasets with Train Versions Only

The classifier was first tested on 3 datasets from the UCI Machine Learning Repository [28]. Recent work [12] has tested some benchmark categorical datasets, including the Wine Recognition database [7], Iris Plants database [6] and the Zoo database [33]. Wine Recognition and Iris Plants have 3 categories, while the Zoo database has 7. These do not have a separate training dataset and are benchmark tests for classifiers. A stopping criterion of 10 iterations was used to terminate the tests. For the Wine dataset, the UCI [28] web page states that the classes are separable, but only RDA [9] has achieved 100% correct classification. Other classifiers achieved: RDA 100%, QDA 99.4%, LDA 98.9%, 1NN 96.1% (z-transformed data) and all results used the leave-one-out technique. So that is the current state-of-the-art. As shown by Table 1, the new classifier can classify to the accuracy required by these benchmark tests. The final column 'Selected Best %' lists the best results found by some other researchers.

| Dataset | Average Error | Best % Error Margin | Correctly Classified | % Correct | Selected Best % |
|---|---|---|---|---|---|
| **Wine** | 0.004 | 25% | 178 from 178 | 100% | 100% |
| **Iris** | 0.005 | 45% | 149 from 150 | 99% | 95.7% |
| **Zoo** | -0.004 | 45% | 101 from 101 | 100% | 94.5% |
| **Abalone** | 0.007 | 49% | 3410 from 4177 | 81% | 73% |
| **Hayes-Roth** | -0.007 | 25% | 131 from 132 | 99% | 50% |
| **Liver** | 0.02 | 35% | 345 from 345 | 100% | 74% |

Table 1. Classifier Test results. Average output error and minimum error margin for the specified number of correct classifications. All datasets points normalised to be in the range 0 to 1. Error margin stopped at 49%.

Three other datasets were tested. These were: the Abalone shellfish dataset [1] with 28 categories and was trained with 20 iterations, or weight transpositions. The Hayes-Roth concept learning dataset [16] with 3 categories, trained to 10 iterations and the BUPA Liver dataset [24], with 2 categories that could be trained in 2 iterations. With the Abalone shellfish dataset, they tried to classify using a decision tree C4.5, a k-NN nearest neighbour and a 1R classifier, from the Weka [30] package. While they reported maybe 73% correct classification, this new method can achieve 81% correct classification.

The paper [20] tested a number of datasets, including Iris, Wine and Zoo, using k-NN and neural network classifiers, with maybe 95.67%, 96% or 94.5% as the best results from one of the classifiers respectively. The values presented here are therefore probably better than that. It also tested the Hayes-Roth dataset, but to only 50% accuracy. Other papers have quoted better results and there is a test dataset available, but without any specified categories. None of the other quoted results are close to 100% however. The paper [11] tested the Liver dataset [24] to 74% accuracy using a sparse grid method, but the new method achieves 100% accuracy in only 2 iterations. The table shows that for all datasets, the error between the desired and the actual output values has reduced to practically zero, but different margins of error are required for the number of correct classifications to be optimised. The percentages still compare favourably with the other researchers' results.

## 6.2   Separate Train and Test Datasets

Four datasets were tried here, where two of them – User Modelling [21] and Bank Notes [25] - were also tested in [12]. They have separate test datasets to the train datasets. This is typically what a supervised neural network should be able to do and the results of this section, given in Table 2, are again favourable. A stopping criterion of 10 iterations was used to terminate the tests.

| Dataset | Average Error | Best % Error Margin | Correctly Classified | % Correct | Selected Best % |
|---|---|---|---|---|---|
| UM | 0.02 | 49% | 143 from 145 | 98.5% | 97.9% |
| Bank notes | -0.05 | 35% | 100 from 100 | 100% | 61% |
| Heart | 0.13 | 35% | 187 from 187 | 100% | 84% |
| Letters | 0.002 | 49% | 3692 from 4000 | 92% | 82% |

Table 2. Classifier Test results. The same criteria as for Table 1, but a separate test dataset to the train dataset.

The User Modelling dataset [21] was used as part of a knowledge-modelling project that produced a new type of classifier in that paper. Their classifier was shown to be much better than the standard ones for the particular problem of web page use, classifying to 97.9% accuracy. This was compared to 85% accuracy for a k-NN classifier and 73.8% for a Bayes classifier. This new model however appears to classify even better, at 98.5% accuracy. Another test tried to classify the bank notes dataset [25]. These were scanned variable values from 'real' or 'fake' banknotes, where the output was therefore binary. This is another different type of problem, where a Wavelet transform might typically be used. The dataset again contained a train and a test dataset, where the best classification realised 100% accuracy. In that paper they quote maybe only 61% correct classification, but other papers have quoted close to 100% correct for similar problems.

A third dataset was a heart classifier from SPECT images [23]. While they noted 84% accuracy on the test dataset using a sparse grid method, the new method can achieve 100% accuracy. A fourth dataset was a letter recognition task [8]. Letters were categorised into one of 26 alphabet types, where there were 20000 instances in total, with 16000 instances in the train set and 4000 instances in the test set. They used a fuzzy exemplar-based rule creation method, but achieved 82% accuracy as compared to 92% accuracy here.

## 7   Conclusions

This paper describes a new type of weight adjustment method that can be used as part of a classifier, or a neural network in particular. It is basically a neural unit with the addition of a very simple rule. The inclusion of the comparison rule however gives the mechanism much more control over weight updates and the unit could still operate in an almost automatic manner. The classifier does not need to learn any complex data rules, but for best results, data normalisation would be required. Another feature is the fact that the weight value can be added or subtracted, and not multiplied, which is the usual mechanism. Another potential advantage is the fact that it can be calculated using only the input and the output values. It is not therefore necessary to fine-tune the classifier with initial weights, or increment/decrement factor amounts, to start with. A stopping criterion should be added however, where each iteration adds a new transposition layer to the matrix. Looking at related work, the learning algorithm is possibly more similar to the Gauss or Pseudo-Newton gradient descent ones [15]. So again, while the method appears to be new, there are similarities with older models. The test results are very surprising. The new classifier appears to work best of all classifiers and across a range of problems. It is also very fast, requiring only a second or less and the setup is really minimal.

Each learning iteration produces a new set of error correction values and so when used, any input value goes through a series of transformations, which is separate for each variable or column value. It is thought that the weight adjustment performs a type of dampening on the error, and so it should reduce for each transposition stage. The orthogonal nature allows the variables to behave slightly differently to each other, where a variable that is close to the desired output value can oscillate around it, while one that is still far away can make larger corrections towards it. There are probably several examples of this type of phenomenon in nature. Another paper that uses an even more orthogonal design is [12], although the results for this paper are maybe slightly better.

**Acknowledgement**

The author wishes to acknowledge an email discussion with Charles Sauerbier of the US Navy, mainly because of its timing. He pointed out a belief that neural networks were a form of cellular automata and several other points, which the author did not fully appreciate, but the simple rule of this paper would push a neural element in that direction. The research itself however derived from a different place, looking at wave shapes and possibly some earlier ideas.

**Addendum**

It has not been made clear in the paper that the classifier actually used the correct output category value to converge to a result when classifying any of the datasets. So even if the classifier had not seen the dataset before, it still used its output category as part of the classification process. This is a major constraint that might be resolved by testing with each output category and selecting the category with the smallest error. However, the average error is also incorrect as it did not take account of negative totals, but it can still be in the hundredths or thousandths after being corrected. So, the results are correct for what is described, apart from the error, but that can still be from a similar scale. A new paper 'An Improved Oscillating-Error Classifier with Branching' has solved the other problems and should also be read.

## References

[1] Asim, A., Li, Y., Xie, Y. and Zhu, Y. (2002). Data Mining For Abalone, Computer Science 4TF3 Project, Supervised by Dr. Jiming Peng, Department of Computing and Software, McMaster University, Hamilton, Ontario.

[2] Brattka, V. (2003). A computable Kolmogorov superposition theorem. Computability and Complexity in Analysis. Informatik Berichte, Vol. 272, pp.7-22.

[3] Carpenter, G., Grossberg, S., and Rosen, D. (1991). Fuzzy ART: Fast stable learning and categorization of analog patterns by an adaptive resonance system. Neural Networks, Vol. 4, pp. 759-771.

[4] Chen, S., Cai, D., Pearce, K., Sun, P.Y-W, Roberts, A.C. and Glanzman, D.L. (2014). Reinstatement of long-term memory following erasure of its behavioral and synaptic expression in Aplysia, eLife 2014;3:e03896, pp. 1 - 21. DOI: 10.7554/eLife.03896.

[5] Dershowitz, N. and Falkovich, E. (2015). Cellular Automata are Generic, U. Dal Lago and R. Harmer (Eds.): Developments in Computational Models 2014 (DCM 2014). EPTCS 179, pp. 17-32, doi:10.4204/EPTCS.179.2.

[6] Fisher, R.A. (1936). The use of multiple measurements in taxonomic problems, Annual Eugenics, 7, Part II, pp. 179-188, also in 'Contributions to Mathematical Statistics' (John Wiley, NY, 1950).

[7] Forina, M. et al. (1991). PARVUS - An Extendible Package for Data Exploration, Classification and Correlation. Institute of Pharmaceutical and Food Analysis and Technologies, Via Brigata Salerno, 16147 Genoa, Italy.

[8] Frey, P.W. and Slate, D.J. (1991). Letter recognition using Holland-style adaptive classifiers, Machine learning, Vol. 6, No. 2, pp. 161-182.

[9] Friedman, J.H. (1989). Regularized Discriminant Analysis, Journal of the American Statistical Association, Vol. 84, No. 405, pp. 165-175.

[10] Gallant, S.I. (1990). Perceptron-Based Learning Algorithms, IEEE Transactions on Neural Networks, Vol. 1, No. 2.

[11] Garcke, J. and Griebel, M., 2002. Classification with sparse grids using simplicial basis functions. Intelligent data analysis, Vol. 6, No. 6, pp. 483-502.

[12] Greer, K. (2015). A Single-Pass Classifier for Categorical Data, available on arXiv at http://arxiv.org/abs/1503.02521.

[13] Greer, K. (2013). Artificial Neuron Modelling Based on Wave Shape, BRAIN. Broad Research in Artificial Intelligence and Neuroscience, Vol. 4, Issues 1-4, pp. 20-25, ISSN 2067-3957 (online), ISSN 2068-0473 (print).

[14] Grossberg, S. (2013). Adaptive resonance theory. Scholarpedia, Vol. 8, No. 5, pp. 1569.

[15] Hagan, M.T. and Menhaj, M.B. (1994). Training Feedforward Networks with the Marquardt Algorithm, IEEE Transactions on Neural Networks, Vol. 5, No. 6, pp. 989-993.

[16]   Hayes-Roth, B. and Hayes-Roth, F. (1977). Concept Learning and the Recognition and Classification of Exemplars, Journal of Verbal Learning and Verbal Behavior, Vol. 16, No. 3, pp. 321-338.

[17]   Hect-Nielsen, R., Neurocomputing, Addison-Wesley, 1990.

[18]   Hinton, G.E., Osindero, S. and Teh, Y.W. (2006). A fast learning algorithm for deep belief nets. Neural computation, Vol. 18, No. 7, pp. 1527-1554.

[19]   Hoffman, J., Tzeng, E., Donahue, J., Jia, Y., Saenko, K. and Darrell, T. (2014). One-Shot Adaptation of Supervised Deep Convolutional Models, arXiv:1312.6204v2 [cs.CV].

[20]   Jiang, Y. and Zhi-Hua Zhou, Z-H. (2004). Editing training data for knn classifiers with neural network ensemble, In Lecture Notes in Computer Science, Vol. 3173, pp. 356-361.

[21]   Kahraman, H.T., Sagiroglu, S. and Colak, I. (2013). The development of intuitive knowledge classifier and the modeling of domain dependent data, Knowledge-Based Systems, Vol. 37, pp. 283-295.

[22]   Kolmogorov, A.N. (1963). On the representation of continuous functions of many variables by superposition of continuous functions of one variable and addition. American Mathematical Society Translation, Vol. 28, No. 2, pp.55-59.

[23]   Kurgan, L.A., Cios, K.J., Tadeusiewicz, R., Ogiela, M. and Goodenday, L.S. (2001). Knowledge Discovery Approach to Automated Cardiac SPECT Diagnosis, Artificial Intelligence in Medicine, Vol. 23, No. 2, pp 149-169.

[24]   Liver dataset (2016). https://archive.ics.uci.edu/ml/datasets/Liver+Disorders.

[25]   Lohweg, V., Dörksen, H., Hoffmann, J. L., Hildebrand, R., Gillich, E., Schaede, J., and Hofmann, J. (2013). Banknote authentication with mobile devices. In IS&T/SPIE Electronic Imaging (pp. 866507-866507). International Society for Optics and Photonics.

[26] Pershin, Y.V., La Fontaine, S. and Di Ventra, M. (2008). Memristive model of amoeba's learning, E-print arXiv:0810.4179, 22 Oct 2008.

[27]   Rojas, R. (1996). Neural Networks: A Systematic Introduction. Springer-Verlag, Berlin and online at books.google.com.

[28]   UCI Machine Learning Repository (2016). http://archive.ics.uci.edu/ml/.

[29]   Waxman, S.G. (2012). Sodium channels, the electrogenisome and the electrogenistat: lessons and questions from the clinic, The Journal of Physiology, pp. 2601 – 2612.

[30]   Weka (2015). http://www.cs.waikato.ac.nz/ml/weka/index.html.

[31]   Widrow, B. and Lehr, M. (1990). 30 Years of adaptive neural networks: perceptron, Madaline and backpropagation, Proc IEEE, Vol. 78, No. 9, pp. 1415-1442.

[32]   Wolfram, S. (1983). Cellular Automata, Los Alamos science.

[33]   Zoo database (2016). https://archive.ics.uci.edu/ml/datasets/Zoo.